\documentclass[11pt,a4paper]{article}
\usepackage[hyperref]{emnlp2020}
\usepackage{times}
\usepackage{latexsym}

\usepackage{booktabs}
\usepackage{url}
\usepackage{tabulary}
\usepackage{graphicx}
\usepackage{amsmath}
\usepackage{amssymb}
\usepackage{bm}
\usepackage{microtype}

\aclfinalcopy 
\def\aclpaperid{1754} 

\title{Graph Convolutions over Constituent Trees for\\ Syntax-Aware Semantic Role Labeling}

\author{ Diego Marcheggiani$^{1}$\thanks{\enspace Research conducted when the author was at the University of Amsterdam.} \hspace{1cm} Ivan Titov$^{2,3}$    \\
    $^1$Amazon \\
 $^2$ILCC, School of Informatics, University of Edinburgh \\
 $^3$ILLC,  University of Amsterdam  \\
    {\tt marchegg@amazon.es} \hspace{0.5cm}  {\tt ititov@inf.ed.ac.uk}}

\date{}

\begin{document}
\maketitle
\begin{abstract}
Semantic role labeling (SRL) is the task of identifying predicates and labeling argument spans with semantic roles. Even though most semantic-role formalisms are built upon constituent syntax, and only syntactic constituents can be labeled as arguments (e.g., FrameNet and PropBank), all the recent work on syntax-aware SRL relies on dependency representations of syntax. In contrast, we show how graph convolutional networks (GCNs) can be used to encode constituent structures and inform an SRL system. Nodes in our SpanGCN correspond to constituents. The computation is done in 3 stages. First, initial node representations are produced by `composing' word representations of the first and last words in the constituent. Second, graph convolutions relying on the constituent tree are performed, yielding syntactically-informed constituent representations. Finally, the constituent representations are `decomposed' back into word representations, which are used as input to the SRL classifier.  
We evaluate SpanGCN against alternatives, including a model using GCNs over dependency trees, and show its effectiveness on standard English SRL benchmarks CoNLL-2005, CoNLL-2012, and FrameNet.
\end{abstract}

\section{Introduction}

\begin{figure}[t]
\begin{center}
\includegraphics[width=1.01\columnwidth]{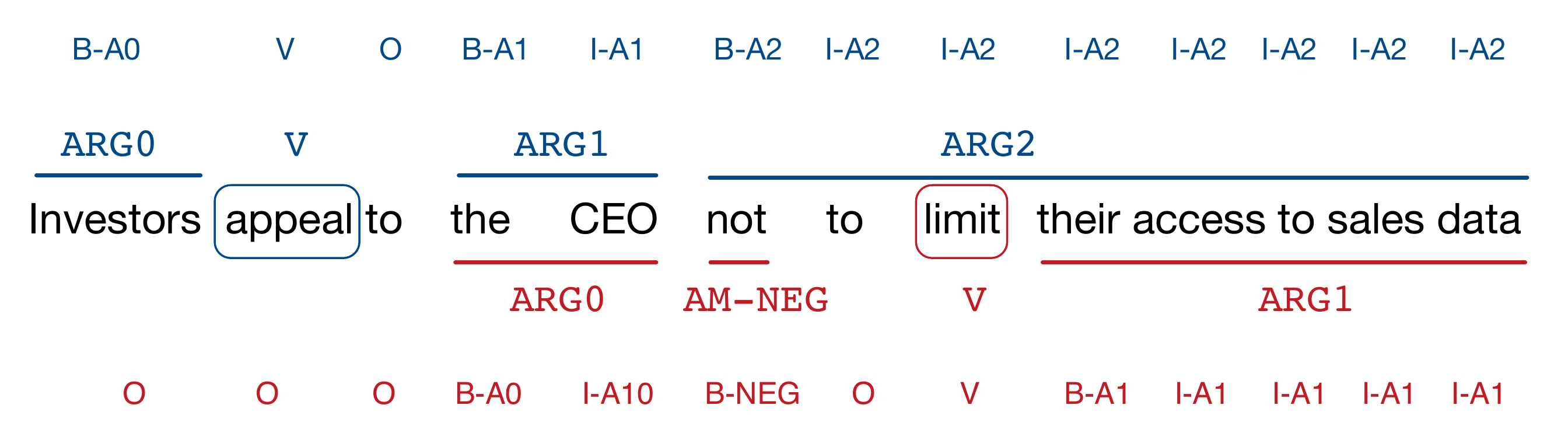}
\vspace{-1ex}
\caption{An example with semantic-role annotation and its reduction to the sequence labeling problem (BIO labels): the argument structure for predicates {\it appeal} and {\it limit} are shown in blue and red, respectively.
\label{fig:propbank_example} 
} 
\vspace{-2ex}
\end{center}
\end{figure}

The task of semantic role labeling (SRL) involves predicting the predicate-argument structure of a sentence. More formally, for every predicate, the SRL model must identify all argument spans and label them with their semantic roles (see Figure \ref{fig:propbank_example}).
The most popular resources for estimating SRL models are PropBank \cite{DBLP:journals/coling/PalmerKG05} and FrameNet \cite{DBLP:conf/acl/BakerFL98}.
In both cases, annotations are made on top of syntactic constituent structures.
Earlier work on SRL hinged on constituent syntactic structure, using the trees to derive features and constraints on role assignments~\cite{gildea2002automatic,DBLP:conf/conll/PradhanHWMJ05,DBLP:journals/coling/PunyakanokRY08}. 
In contrast, modern SRL systems largely ignore treebank syntax \cite{HeLLZ18,HeLLZ17a,DBLP:journals/corr/MarcheggianiFT17,zhou-xu:2015:ACL-IJCNLP} and instead use powerful feature extractors, for example, LSTM sentence encoders.

There have been recent successful attempts to improve neural SRL models using syntax~\cite{DBLP:conf/emnlp/MarcheggianiT17,StrubellVAWM18,DBLP:conf/acl/ZhaoHLB18}.
Nevertheless, they have relied on syntactic dependency representations rather than constituent trees.
In these methods, information from dependency trees is injected into word representations using graph convolutional networks (GCN) \cite{DBLP:journals/corr/KipfW16} or self-attention mechanisms \cite{DBLP:conf/nips/VaswaniSPUJGKP17}.
Since SRL annotations are done on top of syntactic constituents,\footnote{There exists another formulation of SRL, where the focus is on predicting semantic dependency graphs~\cite{DBLP:conf/conll/SurdeanuJMMN08}.
For English, however, these dependency annotations are automatically derived from span-based PropBank.}
we argue that exploiting constituency syntax, rather than dependency one, is more natural and may yield more predictive features for semantic roles.
For example, even though constituent boundaries could be derived from dependency structures, this would require an unbounded number of hops over the dependency structure in GCNs or self-attention. This would be impractical: both \citet{StrubellVAWM18} and \citet{DBLP:conf/emnlp/MarcheggianiT17} use only one hop in their best systems. 

Neural models typically treat SRL as a sequence labeling problem, and hence predictions are made for individual words.
Though injecting dependency syntax into word representations is relatively straightforward, it is less clear how to incorporate constituency syntax.\footnote{Recently, \citet{DBLP:conf/acl/WangJWSW19} proposed different ways of encoding dependency and constituency syntax based on the linearization approaches of \citet{gomez-rodriguez-vilares-2018-constituent}.} 
This work shows how GCNs can be directly applied to span-based structures.
We propose a multi-stage architecture based on GCNs to inject constituency syntax into word representations.
Nodes in our SpanGCN correspond to constituents.
The computation is done in 3 stages. First, initial span representations are produced by `composing' word representations of the first and last words in the constituent. 
Second, graph convolutions relying on the constituent tree are performed, yielding syntactically-informed constituent representations. 
Finally, the constituent representations are `decomposed' back into word representations, which are used as input to the SRL classifier.  
This approach directly injects information about boundaries and syntactic labels of constituents into word representations and also provides information about the word's neighbourhood in the constituent structure.

We show the effectiveness of our approach on three English datasets: CoNLL-2005 \cite{DBLP:conf/conll/CarrerasM05} and CoNLL-2012 \cite{DBLP:conf/conll/PradhanMXUZ12} with PropBank-style \cite{DBLP:journals/coling/PalmerKG05} annotation and on FrameNet 1.5 \cite{DBLP:conf/acl/BakerFL98}~\footnote{Although we tested the model on English datasets, SpanGCN can be applied to constituent trees in any language.}.
By empirically comparing SpanGCN to GCNs over dependency structures, we confirm our intuition that constituents yield more informative features for the SRL task. \footnote{Code available at \url{https://github.com/diegma/span-gcn}.}

SpanGCN may be beneficial in other NLP tasks, where neural sentence encoders are already effective and syntactic structure can provide an additional inductive bias, e.g., logical semantic parsing~\cite{dong2016language} or sentence simplification~\cite{chopra2016abstractive}. 
Moreover, in principle, SpanGCN can be applied to other forms of span-based linguistic representations (e.g., co-reference, entity+relations graphs, semantic and discourse structures).
However, we leave this for future work.

\section{Constituency Tree Encoding}
\label{sec:cons_encoding}
\begin{figure}[t]
\begin{center}
\includegraphics[width=1.05\columnwidth]{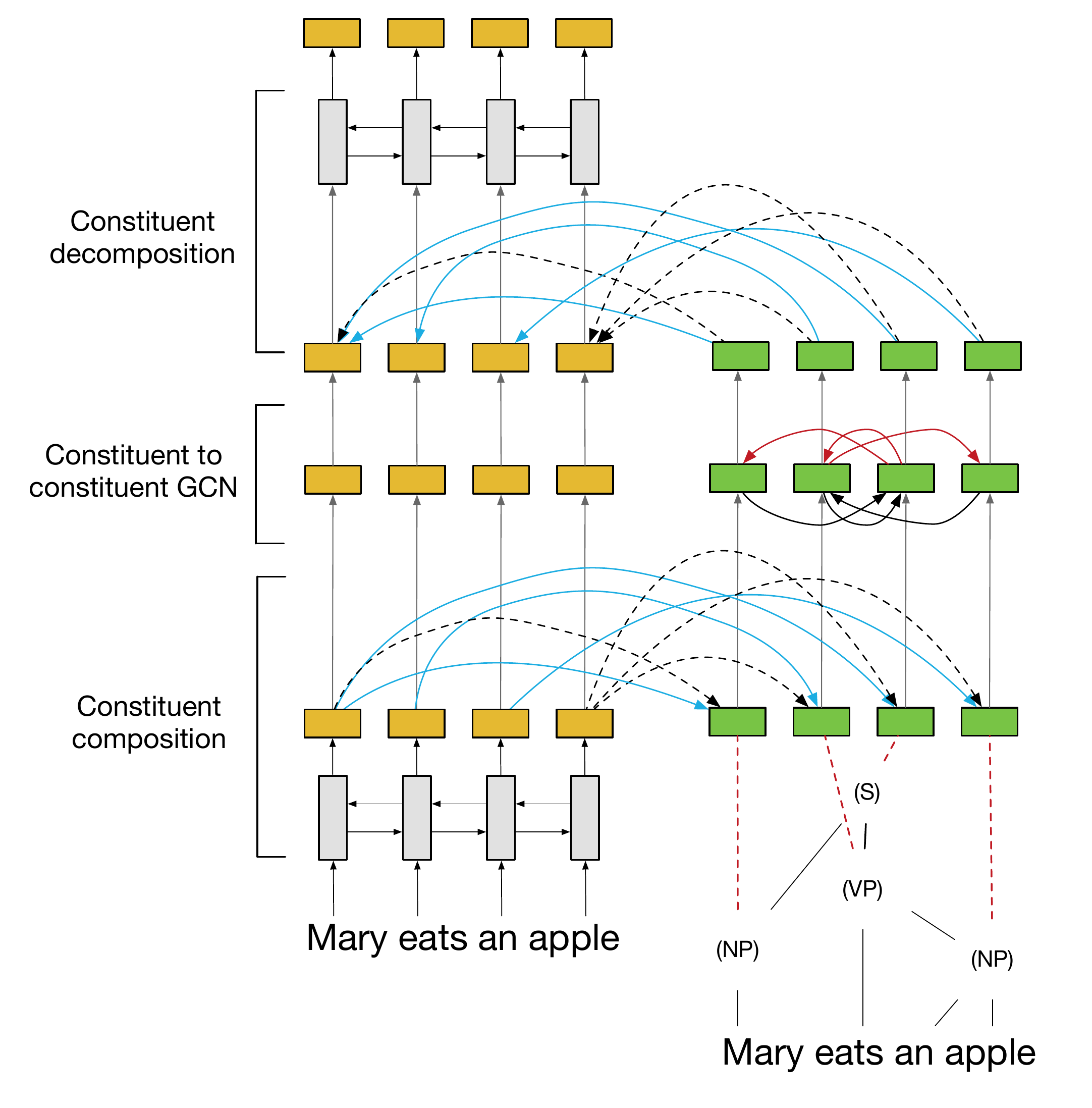}

\vspace{-1ex}
\caption{SpanGCN encoder. 
First, for each constituent, an initial representation is produced by composing the start and end tokens' BiLSTM states (cyan and black dashed arrows, respectively). This is followed by a constituent GCN: red and black arrows represent parent-to-children and children-to-parent messages, respectively. Finally, the constituent is decomposed back: each constituent sends messages to its start and end tokens.  
\label{fig:model} 
} 
\vspace{-2ex}
\end{center}
\end{figure}

The architecture for encoding constituency trees uses two building blocks, a bidirectional LSTM for encoding sequences and a graph convolutional network for encoding graph structures.

\subsection{BiLSTM encoder}
A bidirectional LSTM (BiLSTM) \cite{DBLP:journals/corr/Graves13} consists of two LSTMs \cite{DBLP:journals/neco/HochreiterS97}, one that encodes the left context of a word and one that encodes the right context. 
In this paper, we use alternating-stack BiLSTMs as introduced by \citet{zhou-xu:2015:ACL-IJCNLP}, where the forward LSTM is used as input to the backward LSTM. 
As in \citet{HeLLZ17a}, we employ highway connections \cite{DBLP:conf/nips/SrivastavaGS15} between layers and recurrent dropout \cite{DBLP:conf/nips/GalG16} to avoid overfitting.

\subsection{GCN}
The second building block we use is a graph convolutional network \cite{DBLP:journals/corr/KipfW16}.
GCNs are neural networks that, given a graph, compute the representation of a node conditioned on the neighboring nodes.
It can be seen as a message-passing algorithm where a node's representation is updated based on `messages' sent by its neighboring nodes \cite{gilmer2017neural}.

The input to GCN is an undirected graph $\mathcal{G} =(\mathcal{V}, \mathcal{E})$, where $\mathcal{V}$ ($|V|=n$) and $\mathcal{E}$ are sets of nodes and edges, respectively. \newcite{DBLP:journals/corr/KipfW16} assume that the set of edges $\mathcal{E}$ contains also a self-loop, i.e., $(v, v) \in \mathcal{E}$ for any $v$.
We refer to the initial representation of nodes with a matrix $X \in \mathbb{R}^{m \times n}$, with each of its column $x_v \in \mathbb{R}^m$ ($v \in \mathcal{V}$) encoding node features.
The new node representation is computed as
\begin{equation*}
\label{eq:gcns}
h^{}_v = ReLU\left(\sum_{u \in \mathcal{N}(v)} ( U^{} x_{u} + b) \right),
\end{equation*}
where $U \in \mathbb{R}^{m \times m}$ and $b\in \mathbb{R}^{m}$ are a weight matrix and a bias, respectively; $\mathcal{N}(v)$ are neighbors of $v$; 
$ReLU$ is the rectifier linear unit activation function. 

The original GCN definition assumes that edges are undirected and unlabeled. We take inspiration
from dependency GCNs \cite{DBLP:conf/emnlp/MarcheggianiT17} introduced for dependency syntactic structures.
Our update function is defined as 
\begin{align}
\nonumber
\label{eq:const_gcn}
h_v^{'} = & ReLU\Large( LayerNorm\Large( \\
& \sum_{u \in \mathcal{N}(v)} g_{v, u}(U^{}_{T_c(u,v)} h_{u} + b^{}_{T_f(u,v)}) \Large)\Large),
\end{align}
where $LayerNorm$ refers to layer normalization \cite{DBLP:journals/corr/BaKH16} applied after summing the messages.
Expressions $T_f(u,v)$ and $T_c(u,v)$ are fine-grained and coarse-grained versions of edge labels. For example, $T_c(u,v)$ may simply return the direction of an arc (i.e. whether the message flows along the graph edge or in the opposite direction), whereas the bias can provide some additional syntactic information. The typing decides how many parameters GCN has. It is crucial to keep the number of coarse-grained types low as the model will have to estimate one 
$\mathbb{R}^{m \times m}$ matrix per coarse-grained type. 
We formally define the types in the next section. 
We used scalar gates $g_{u, v}$ to weight the contribution of each node in the neighborhood and potentially ignore irrelevant edges:
\begin{equation}
g_{u, v}^{} = \sigma\left( \hat{u}^{}_{T_c(u,v)} \cdot h_u^{} + \hat{b}_{T_f(u,v)}\right),
\end{equation}
where $ \sigma$ is the sigmoid activation function, whereas $\hat{u}^{}_{T_c(u,v)} \in \mathbb{R}^m$ and $\hat{b}_{T_f(u,v)} \in \mathbb{R}$
are edge-type-specific parameters.

Now, we show how to compose GCN and LSTM layers to produce a syntactically-informed encoder.

\subsection{SpanGCN}
\label{sec:message_passing}
Our model is shown in Figure \ref{fig:model}. It is composed of three modules: constituent composition, constituent GCN, and constituent decomposition. Note that there is no parameter sharing across these components.

\paragraph{Constituent composition}
The model takes as input word representations which can either be static word embeddings or contextual word vectors \cite{DBLP:conf/naacl/PetersNIGCLZ18,DBLP:journals/corr/abs-1907-11692,DBLP:conf/naacl/DevlinCLT19}.
The sentence is first encoded with a BiLSTM to obtain a context-aware representation of each word.
A constituency tree is composed of words ($ \mathcal{V}_{w}$) and constituents ($\mathcal{V}_{c}$).\footnote{We slightly abuse the notation by referring to non-terminals as constituents: part-of-speech tags (typically `pre-terminals') are stripped off from our trees.}
We add representations (initially zero vectors) for each constituent in the tree; they are shown as green blocks in Figure \ref{fig:model}. 
Each constituent representation is computed using GCN updates (Equation \ref{eq:const_gcn}), relying on the word representation corresponding to the beginning of its span and the representation corresponding to the end of its span. The coarse-grained types $T_c(u,v)$ here are binary, distinguishing messages from start tokens vs. end tokens. 
The fine-grained edge types $T_f(u,v)$ encode additionally the constituent label (e.g., NP or VP).

\paragraph{Constituent GCN}
The constituent composition stage is followed by a layer where constituent nodes exchange messages.
This layer makes sure that information about children gets incorporated into representations of immediate parents and vice versa. GCN operates on the graph with nodes corresponding to all constituents ($\mathcal{V}_{c}$) in the trees. The edges connect constituents and their immediate children in the syntactic tree and do it in both directions. Again, the updates are defined as in Equation~\ref{eq:const_gcn}. As before, $T_c(u,v)$ is binary, now distinguishing parent-to-children messages from
children-to-parent messages. $T_f(u,v)$ additionally includes the label of the constituent sending the message. For example, consider the computation of the VP constituent in Figure~\ref{fig:model}. It receives a message from the $S$ constituent, this is a parent-to-child message, and the `sender' is $S$;
these two factors determine $T_f(u,v)$ and, as a result, the parameters used in computing the corresponding message.

\paragraph{Constituent decomposition}
At this point, we 'infuse' words with information coming from constituents. The graph here is the inverse of the one used in the composition stage: the constituents pass the information to the first and the last words in their spans.
As in the composition stage, $T_c(u,v)$ is binary, distinguishing messages to start and end tokens. The fine-grained edge types, as before, additionally include the constituent label. 
To spread syntactic information across the sentence, we use a further BiLSTM layer.

Note that residual connections indicated in grey in Figure~\ref{fig:model}, let the model bypass GCN if / where needed.

\section{Semantic Role Labeling}
SRL can be cast as a sequence labeling problem where given an input sentence $\mathbf{x}$ of length $T$, and the position of the predicate in the sentence $p \in T$, the goal is to predict a BIO sequence of semantic roles $\mathbf{y}$ (see Figure~\ref{fig:propbank_example}).
We test our model on two different SRL formalisms, PropBank and FrameNet.
\paragraph{PropBank}
In PropBank conventions, a frame is specific to a predicate sense. For example, for the predicate {\it make},
it distinguishes `make.01' (`create') frame from `make.02' (`cause to be') frame.
Though roles are formally frame-specific (e.g., {\it A0} is the `creator' for the frame `make.01' and the `writer' for the frame `write.01'), there are certain cross-frame regularities. For example, {\it A0} and {\it A1} tend to correspond to proto-agents and proto-patients, respectively.

\paragraph{FrameNet}
In FrameNet, every frame has its own set of role labels (frame elements in FrameNet terminology).\footnote{Cross-frame relations (e.g., the frame hierarchy) present in FrameNet can, in principle, be used to establish correspondences between a subset of roles. }
This makes the problem of predicting role labels harder.
Differently from PropBank, lexically distinct predicates (lexical units or targets in FrameNet terms) may evoke the same frame. For example, {\it need} and {\it require} both can trigger frame `Needing'.

As in previous work we compare to, we assume to have access to gold frames \cite{DBLP:conf/emnlp/SwayamdiptaTLZD18,YangM17}. 

\section{Semantic Role Labeling Model}
\label{sec:model}
For both PropBank and FrameNet, we use the same model architecture.
\paragraph{Word representation} We represent words with
pretrained word embeddings,
and we keep them fixed during training. Word embeddings are concatenated with 100-dimensional embeddings of a predicate binary feature (indicating if the word is the target predicate or not).
Before concatenation, the pretrained embeddings are passed through layer normalization \cite{DBLP:journals/corr/BaKH16} and dropout \cite{DBLP:journals/jmlr/SrivastavaHKSS14}.
Formally,
\begin{equation*}
x_{t} = dropout(LayerNorm(w_t)) \circ predemb(t)),
\end{equation*}
where $predemb(t)$ is a function that returns the embedding for the presence or absence of the predicate at position $t$.
The obtained embedding $x_{t}$ is then fed to the sentence encoder.

\paragraph{Sentence encoder}
As a sentence encoder we use SpanGCN introduced in Section \ref{sec:cons_encoding}. 
SpanGCN is fed with word representations $x_{t}$. 
Its output is a sequence of hidden vectors that encode syntactic information for each candidate argument~$h_{t}$.
As a baseline, we use a syntax-agnostic sentence encoder that is the reimplementation of the encoder of \citet{HeLLZ17a} with stacked alternating LSTMs, i.e., our model with the three GCN layers stripped off.\footnote{To have a fair baseline, we independently tuned the number of BiLSTM layers for our model and the baseline.} 

\paragraph{Bilinear scorer}
Following \citet{StrubellVAWM18}, we used a bilinear scorer:
\begin{equation*}
s_{pt} = (h^{pred}_{p})^{T} U (h^{arg}_{t}).
\end{equation*}
$h^{pred}_{p}$ and $h^{role}_{t}$ are a non-linear projection of the predicate $h_{p}$ at position $p$ in the sentence and the candidate argument $h_{t}$.
The scores $s_{pt}$ are passed through the softmax function and fed to the conditional random field (CRF) layer.

\paragraph{Conditional random field} For the output layer, we use a first-order Markov CRF \cite{DBLP:conf/icml/LaffertyMP01}. 
We use the Viterbi algorithm to predict the most likely label assignment at testing time.
At training time, we learn the scores for transitions between BIO labels.
The entire model is trained to minimize the negative conditional log-likelihood:
\begin{equation*}
\mathcal{L} = - \sum_{j}^{N} \log P(\mathbf{y}|\mathbf{x}, p)
\end{equation*}
where $p$ is the predicate position for the training example $j$.

\section{Experiments}

\subsection{Data and setting}

We experiment on the CoNLL-2005 and CoNLL-2012 (OntoNotes) datasets and use the CoNLL 2005 evaluation script for evaluation.
We also apply our approach to FrameNet 1.5 with the data split of \citet{DBLP:journals/coling/DasCMSS14} and follow the official evaluation set-up from the SemEval’07 Task 19 on frame-semantic parsing \cite{DBLP:conf/semeval/BakerEE07}.

We train the self-attentive constituency parser of \citet{DBLP:conf/acl/KleinK18}\footnote{https://github.com/nikitakit/self-attentive-parser} on the training data of the CoNLL-2005 dataset (Penn Treebank) and parse the development and test sets of CoNLL-2005 dataset. 
We apply the same procedure for the CoNLL-2012 dataset.
We perform 10-fold jackknifing to obtain syntactic predictions for the training set of CoNLL-2005 and CoNLL-2012.
For FrameNet, we parse the entire corpus with the parser trained on the training set of CoNLL-2005.
All hyperparameters are reported in Appendix \ref{sec:implementation_details}.

\subsection{Importance of syntax and ablations}

\begin{table}
\begin{tabular}{llll}
& \multicolumn{3}{c}{Dev} \\ \cline{2-4} 
 & P & R & F1 \\ \hline \hline
Baseline & 82.78 & 83.58 & 83.18 \\
\hline 
SpanGCN & 84.48 & 84.26 & 84.37 \\
\thinspace \thinspace \thinspace \thinspace (w/o BiLSTM) & 83.31 & 83.35 & 83.33 \\
\hline
SpanGCN \small{(Gold)} & 90.50 & 90.65 & 90.58 \\
\thinspace \thinspace \thinspace \thinspace (w/o BiLSTM) & 88.96 & 90.02 & 89.49 \\
\hline 
DepGCN & 83.4 & 83.73 & 83.56 \\
\thinspace \thinspace \thinspace \thinspace (w/o BiLSTM) & 83.01 & 83.18 & 83.09 \\
\hline 
\end{tabular}
\caption{Results with predicted and gold syntax on the CoNLL-2005 development set. \label{tab:gold_syntax}}
\end{table}

\begin{figure}[t]
\begin{center}
\includegraphics[width=1.00\columnwidth]{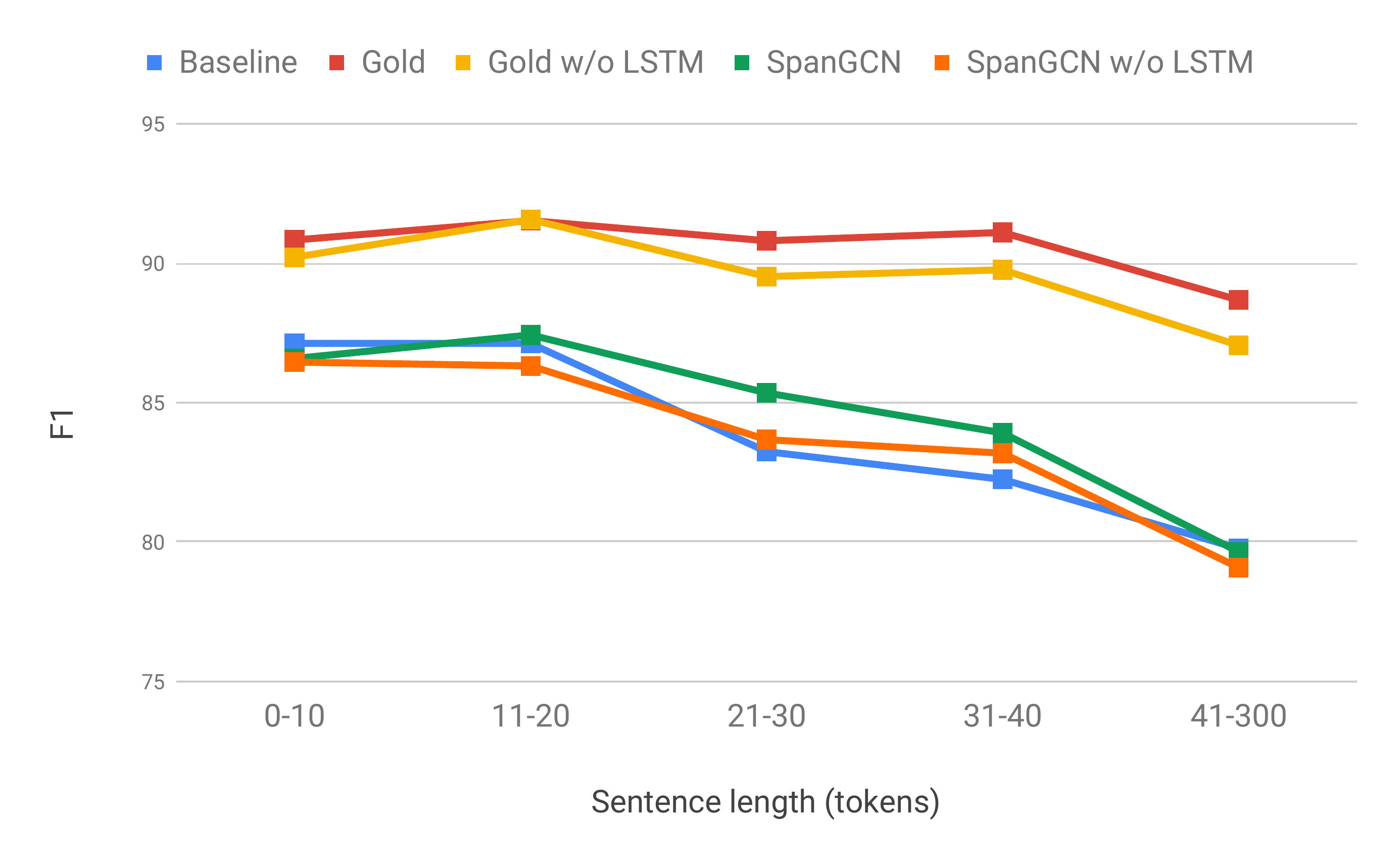}
\includegraphics[width=1.00\columnwidth]{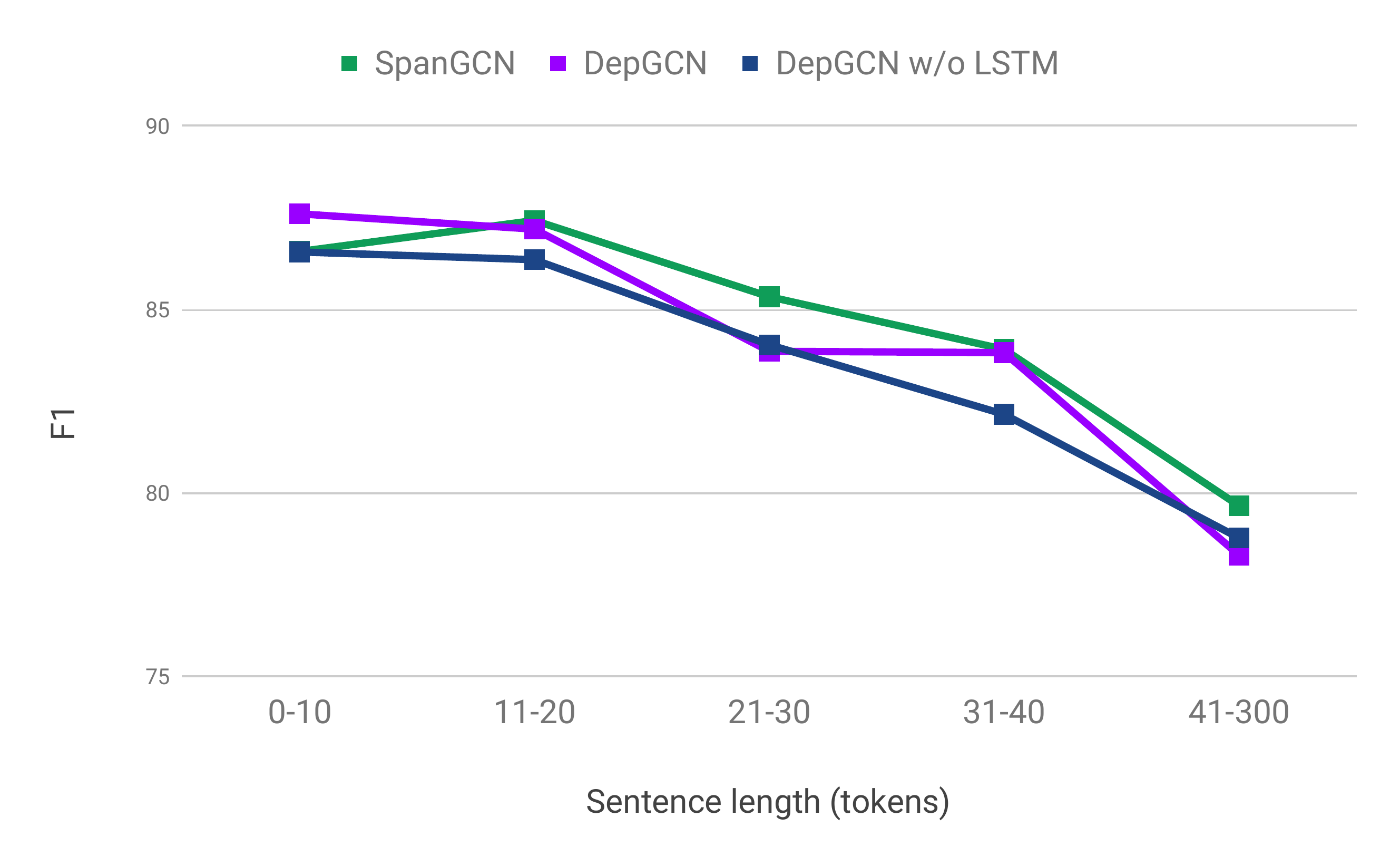}
\caption{CoNLL-2005 F1 score as a function of sentence length.\label{fig:sent_len} 
} 
\end{center}
\end{figure}

\begin{figure}[t]
\begin{center}
\includegraphics[width=1.00\columnwidth]{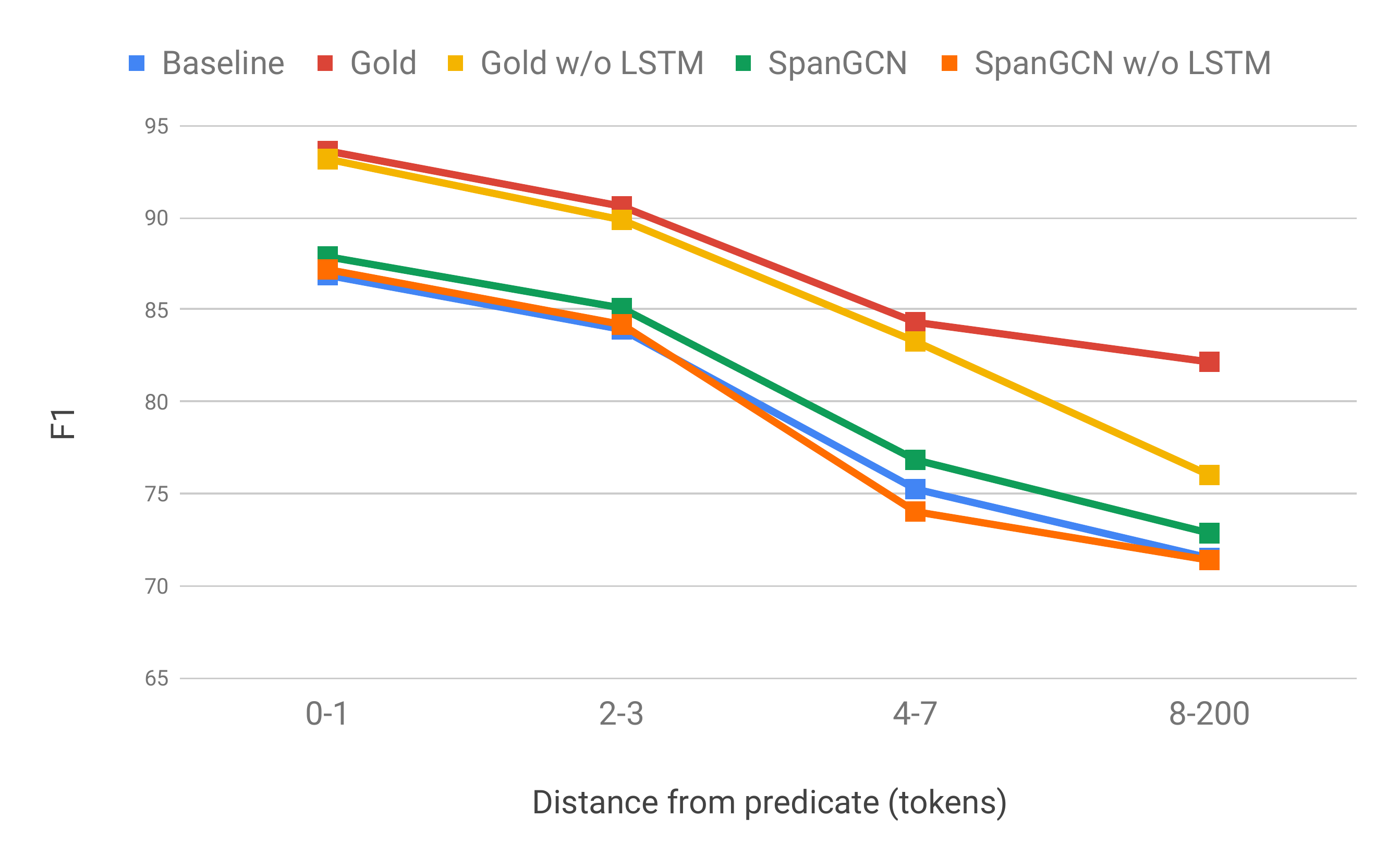}
\includegraphics[width=1.00\columnwidth]{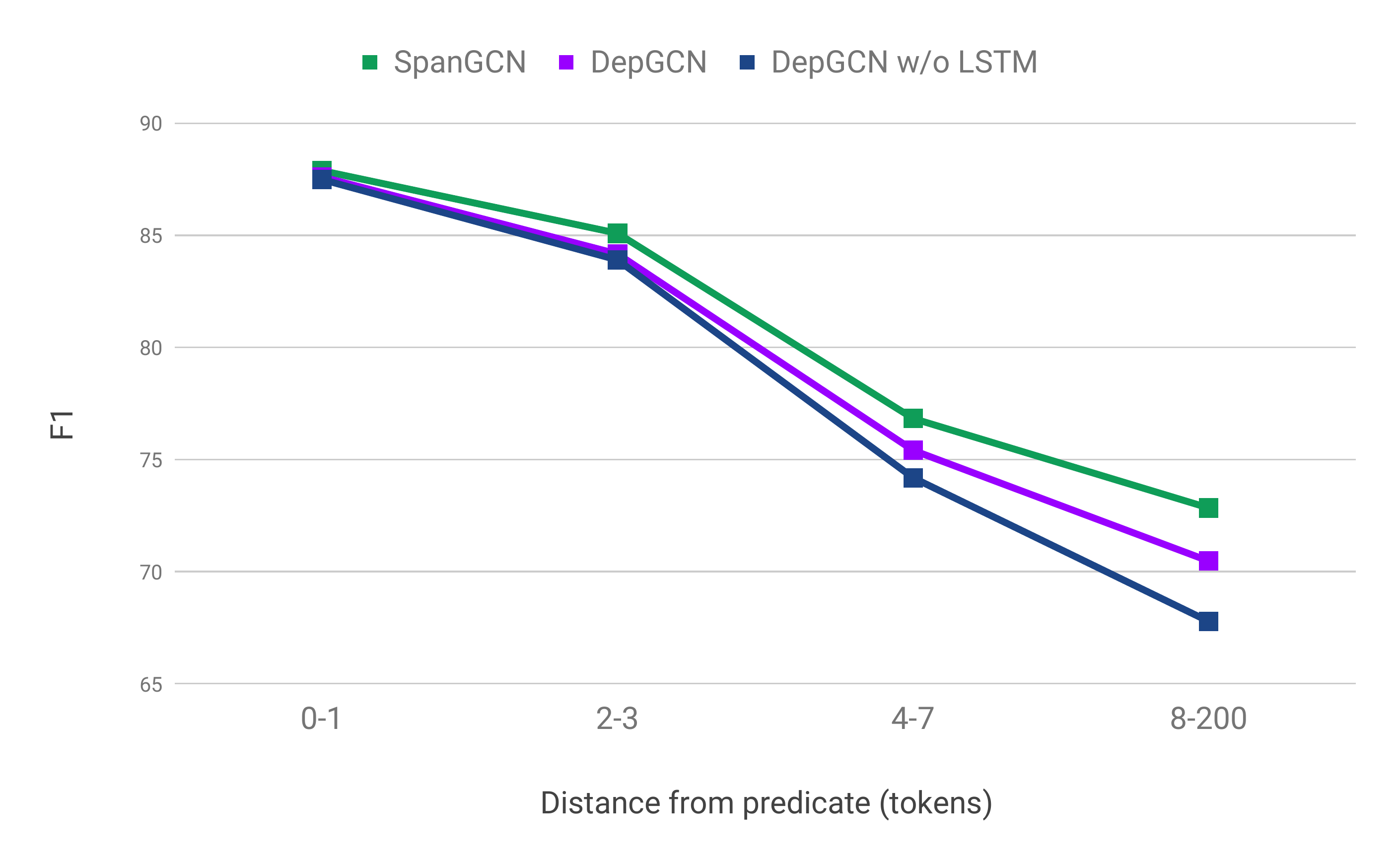}
\caption{CoNLL-2005 F1 score as a function of the distance of a predicate from its arguments.\label{fig:pred_distance} 
} 
\end{center}
\end{figure}

\begin{figure}[t]
\begin{center}
\includegraphics[width=1.00\columnwidth]{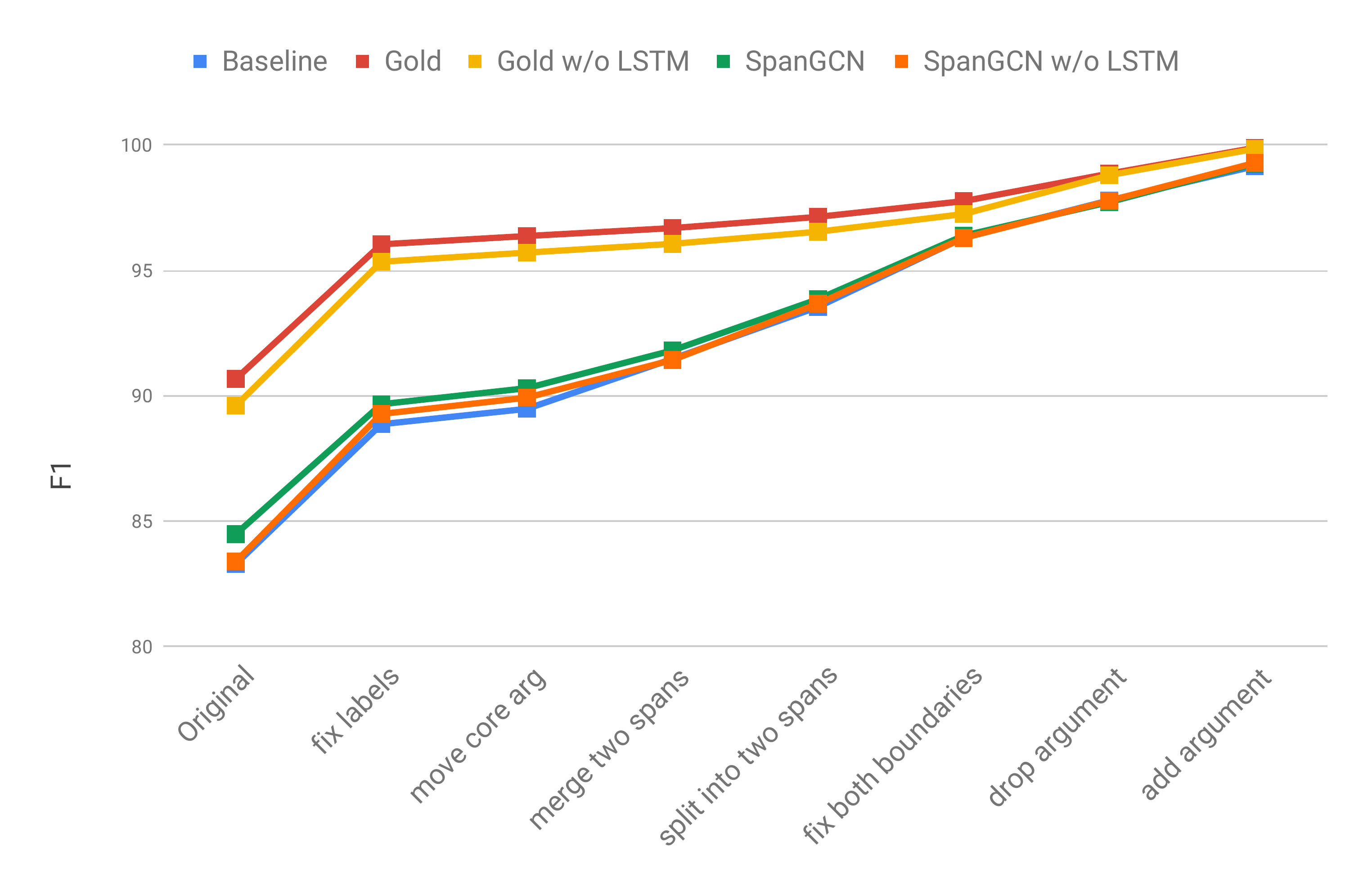}
\includegraphics[width=1.00\columnwidth]{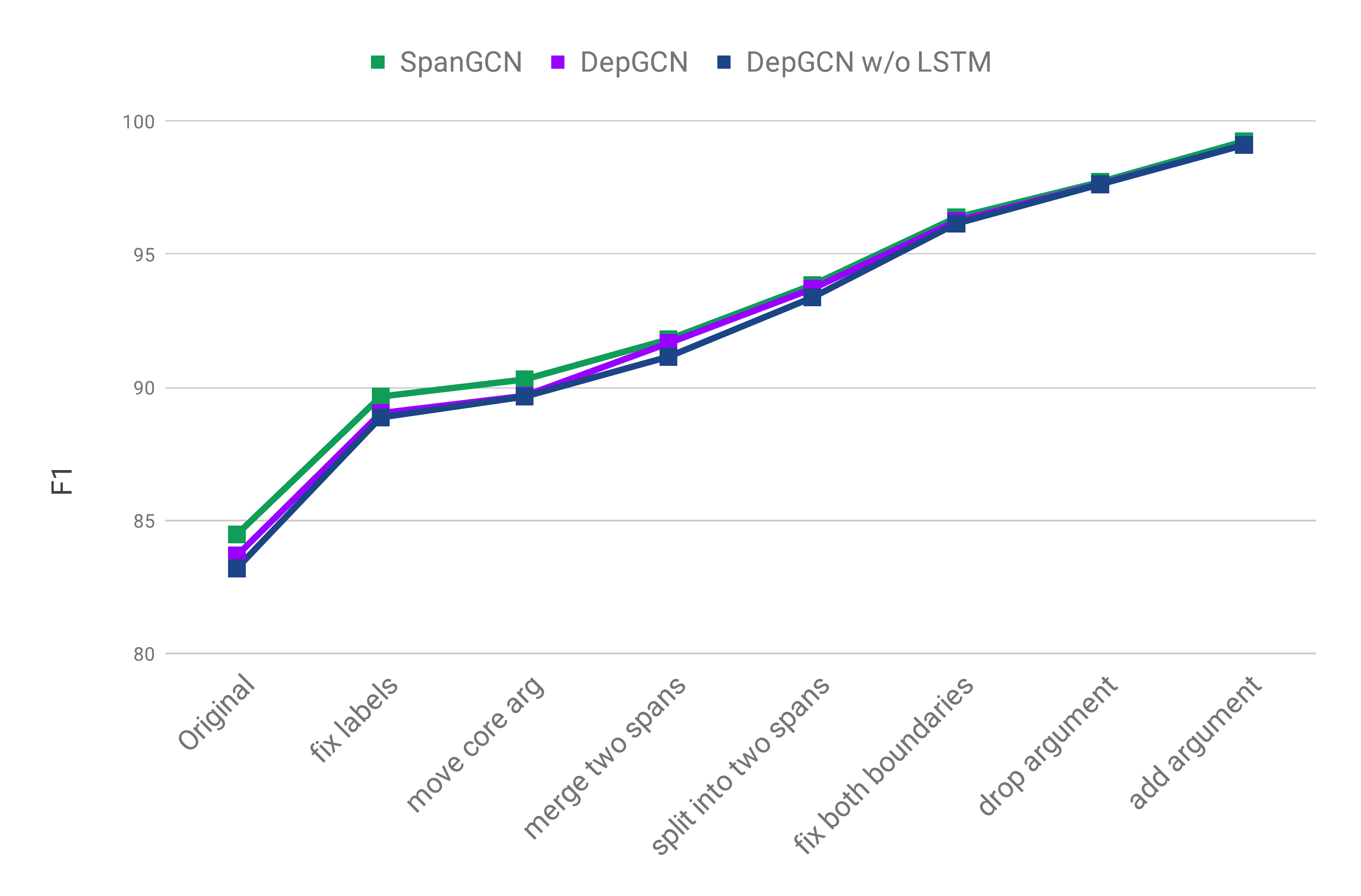}
\caption{Performance of CoNLL-2005 models after performing corrections from \citet{HeLLZ17a}.\label{fig:corrections} 
} 
\end{center}
\end{figure}

Before comparing our full model to state-of-the-art SRL systems, we show that our model genuinely benefits from incorporating syntactic information and motivate other modeling decisions (e.g., the presence of BiLSTM layers at the top). 

We perform this analysis on the CoNLL-2005 dataset. 
We also experiment with gold-standard syntax, as this provides an upper bound on what
SpanGCN can gain from using syntactic information.

\begin{table}[ht]
\scalebox{0.9}{
\begin{tabular}{llllll}
& \multicolumn{3}{c}{WSJ Test} \\ \cline{2-4}
 Single & $\bm{P}$ & $\bm{R}$ & $\bm{F_1}$ \\ \hline \hline
\citet{HeLLZ17a} & 83.1 & 83.0 & 83.1 \\ 
\citet{HeLLZ18} & 84.2 & 83.7 & 83.9 \\ 
\citet{TanWXCS18} & 84.5 & 85.2 & 84.8 \\
\citet{DBLP:conf/emnlp/OuchiS018} & 84.7 & 82.3 & 83.5 \\
 \citet{StrubellVAWM18}\small{(LISA)}$\dagger\ddagger$ & 84.7& 84.6& 84.6 \\ 
 \hline
SpanGCN$\dagger$ & 85.8 & 85.1 & {85.4} \\ 
\hline 

 & & & \\
 
Single / Context. Emb. & & & \\ \hline \hline
\citet{HeLLZ18}{\small (ELMo)} & - & - & 87.4 \\ 
\citet{DBLP:journals/corr/abs-1901-05280}{\small (ELMo)} & 87.9 & 87.5 & 87.7 \\ 
\citet{DBLP:conf/emnlp/OuchiS018}{\small (ELMo)} & 88.2 & 87.0 & 87.6\\
\citet{DBLP:conf/acl/WangJWSW19}{\small (ELMo)}$\dagger$ & - & - & 88.2 \\
\hline 
SpanGCN {\small (ELMo)}$\dagger$ & 87.5 & 87.9 & 87.7 \\
SpanGCN {\small (RoBERTa)}$\dagger$ & 87.7 & 88.1 & 87.9 \\
\hline

 & & & \\
& \multicolumn{3}{c}{Brown Test} \\ \cline{2-4}
 Single & $\bm{P}$ & $\bm{R}$ & $\bm{F_1}$\\ \hline \hline
\citet{HeLLZ17a} & 72.9 & 71.4 & 72.1\\ 
\citet{HeLLZ18} & 74.2 & 73.1 & 73.7\\ 
\citet{TanWXCS18} & 73.5 & 74.6 & 74.1\\
\citet{DBLP:conf/emnlp/OuchiS018} & 76.0& 70.4 & 73.1\\
\citet{StrubellVAWM18}\small{(LISA)}$\dagger\ddagger$& 74.8 &74.3& 74.6\\
\hline
SpanGCN$\dagger$ & 76.2 & 74.7 & 75.5 \\ 
\hline

 & & & \\

Single / Context. Emb. & & & \\ \hline \hline
\citet{HeLLZ18}{\small (ELMo)} & - & - & 80.4\\ 

\citet{DBLP:journals/corr/abs-1901-05280}{\small (ELMo)} & 80.6 & 80.4 & 80.5 \\ 
\citet{DBLP:conf/emnlp/OuchiS018}{\small (ELMo)} & 79.9 & 77.5 & 78.7\\
\citet{DBLP:conf/acl/WangJWSW19}{\small (ELMo)}$\dagger$ & - & - & 79.3 \\
\hline 
SpanGCN{\small (ELMo)}$\dagger$ &  79.4 & 79.6 & 79.5 \\
SpanGCN{\small (RoBERTa)}$\dagger$ &  80.5 & 80.7 & 80.6 \\
\hline

\end{tabular}
}
\caption{Precision, recall and $\bm{F_1}$ on the CoNLL-2005 test sets. $\dagger$ indicates syntactic models and $\ddagger$ indicates multi-task learning models. \label{tab:conll05-results}}
\end{table}

From Table \ref{tab:gold_syntax}, we can see that SpanGCN improves over the syntax-agnostic baseline by $1.2$\% F1, a substantial boost from using predicted syntax. We can also observe that it is important to have the top BiLSTM layer.
When we remove the BiLSTM layer, the performance drops by 1\% F1.
Interestingly, without this last layer, SpanGCN's performance is roughly the same as that of the baseline.
This shows the importance of spreading syntactic information from constituent boundaries to the rest of the sentence.

When we provide to SpanGCN gold-standard syntax instead of the predicted one, the SRL scores improve greatly.\footnote{The syntactic parser we use scores 92.5\% F1 on the development set.} 
This suggests that, despite its simplicity (e.g., somewhat impoverished parameterization of constituent GCNs), SpanGCN
is capable of extracting predictive features from syntactic structures.

We also measure the performance of the models above as a function of sentence length (Figure \ref{fig:sent_len}), and as a function of the distance between a predicate and its arguments (Figure \ref{fig:pred_distance}).
\begin{table}
\center
\scalebox{0.90}{
\begin{tabular}{llll}
& \multicolumn{3}{c}{Test} \\ 
\cline{2-4} 
Single & $\bm{P}$ & $\bm{R}$ & $\bm{F_1}$ \\ \hline \hline
\citet{HeLLZ17a} & 81.7 & 81.6 & 81.7 \\ 
\citet{HeLLZ18} & - & - & 82.1 \\ 
\citet{TanWXCS18} & 81.9 & 83.6 & 82.7 \\
\citet{DBLP:conf/emnlp/OuchiS018} & 84.4 & 81.7 & 83.0 \\
\citet{DBLP:conf/emnlp/SwayamdiptaTLZD18}$\dagger\ddagger$ & 85.1 & 81.2 & 83.8 \\

\hline 
SpanGCN$\dagger$ & 84.5 & 84.3 & 84.4 \\ 
\hline 

 & & & \\

Single / Context. Emb. & & & \\ \hline \hline
\citet{DBLP:conf/naacl/PetersNIGCLZ18}{\small (ELMo)} & - & - & 84.6 \\
\citet{HeLLZ18}{\small (ELMo)} & - & - & 85.5 \\ 
\citet{DBLP:journals/corr/abs-1901-05280}{\small (ELMo)} & 85.7 & 86.3 & 86.0 \\ 
\citet{DBLP:conf/emnlp/OuchiS018}{\small (ELMo)} & 87.1 & 85.3 & 86.2 \\

\citet{DBLP:conf/acl/WangJWSW19}{\small (ELMo)}$\dagger$ & - & - & 86.4 \\
\hline 
SpanGCN {\small (ELMo)}$\dagger$ & 86.3 & 86.8 & 86.5 \\
SpanGCN {\small (RoBERTa)}$\dagger$ & 86.5 & 87.1 & 86.8 \\
\hline 
\end{tabular}
}
\caption{Precision, recall and $\bm{F_1}$ on the CoNLL-2012 test set. $\dagger$ indicates syntactic models and $\ddagger$ indicates multi-task learning models. \label{tab:conll12-results}}
\end{table}
Not surprisingly, the performance of every model degrades with the length.
For the model using gold syntax, the difference between F1 scores on short sentences and long sentences is smaller (2.2\% F1) than for the models using predicted syntax (6.9\% F1). This is also expected as in the gold-syntax set-up, SpanGCN can rely on perfect syntactic parses even for long sentences. In contrast, in the realistic set-up syntactic features start to be unreliable.
SpanGCN performs on par with the baseline for very short and very long sentences.
Intuitively, for short sentences, BiLSTMs may already encode enough syntactic information, while for longer sentences, the quality of predicted syntax is not good enough to get gains over the BiLSTM baseline.
When considering the performance of each model as a function of the distance between a predicate and its arguments, 
we observe that all models struggle with more `remote' arguments. 
Evaluated in this setting, SpanGCN is slightly better than the baseline.

We also check what kind of errors these models make by using an oracle to correct one error type at the time and measuring the influence on the performance \cite{HeLLZ17a}.
Figure \ref{fig:corrections} (top) shows the results.
We can see that all the models make the same fraction of mistakes in labeling arguments, even with gold syntax.
It is also clear that using gold syntax and, to a lesser extent, predicted syntax, helps the model to figure out the exact boundaries of argument spans. These results
also suggest that using gold-syntax leads to many fewer span-related errors: fixing these errors (merge two spans, spit into two spans, fix both boundaries)
yields 6.1\% and 1.4\% improvements, when using predicted and gold syntax, respectively.
The BiLSTM is even weaker here (6.8\% increase in F1).

\paragraph{SpanGCN vs. DependencyGCN} To show the benefits of using constituency syntax, we compare SpanGCN with the dependency GCN (DepGCN) of \citet{DBLP:conf/emnlp/MarcheggianiT17}. 
We use DepGCN in our architecture in place of the 3-stage SpanGCN. 
We obtain dependency trees by transforming the predicted constituency trees with CoreNLP  \cite{DBLP:conf/coling/MarneffeM08}.
Table \ref{tab:gold_syntax} shows that while the model with DepGCN preforms +0.38\% better than the baseline, it performs worse than SpanGCN 83.56 vs. 84.36 F1. 
In Figure~\ref{fig:sent_len} (bottom), we also compare the performance of the two syntactic encoders as a function of sentence length. 
Interestingly, DepGCN performs slightly better than SpanGCN on short sentences.
Figure~\ref{fig:pred_distance} (bottom) shows that SpanGCN performs on par with DepGCN when arguments are close to the predicate but better for more distant arguments.
As with SpanGCN, Figure \ref{fig:sent_len} and \ref{fig:pred_distance} (bottom) show that
adding a BiLSTM on top of DepGCN helps to capture long range dependencies. 
In Figure \ref{fig:corrections}(bottom), we show the different behaviour of DepGCN with respect to SpanGCN in terms of prediction mistakes. 
Unsurprisingly, the main mistake that DepGCN makes is on deciding span boundaries. Fixing span related errors (merge two spans, spit into two spans, fix both boundaries) yields an improvement of 6.6\% for DepGCN vs. 6.1\% of SpanGCN.

\subsection{Comparing to the state of the art}

We compare SpanGCN with state-of-the-art models on both CoNLL-2005 and CoNLL-2012.\footnote{We only consider single, non-ensemble models.}
\paragraph{CoNLL-2005} In Table \ref{tab:conll05-results} (Single) we show results on the CoNLL-2005 dataset. 
We compare the model with approaches that use syntax \cite{StrubellVAWM18,DBLP:conf/acl/WangJWSW19} and with syntax-agnostic models \cite{HeLLZ18,HeLLZ17a,TanWXCS18,DBLP:conf/emnlp/OuchiS018}.
SpanGCN obtains the best results also outperforming the multi-task self-attention model of \citet{StrubellVAWM18}\footnote{We compare with the LISA model where no ELMo information \cite{DBLP:conf/naacl/PetersNIGCLZ18} is used, neither in the syntactic parser nor the SRL components.} on the WSJ (in-domain) (85.43 vs. 84.64 F1) and Brown (out-of-domain) (75.45 vs. 74.55 F1) test sets.
The performance on the Brown test shows that SpanGCN is robust with nosier syntax.

\paragraph{CoNLL-2012}In Table \ref{tab:conll12-results} (Single) we report results on the CoNLL-2012 dataset.
SpanGCN obtains 84.4 F1, outperforming all previous models evaluated on this data.

\paragraph{Experiments using contextualized embeddings}
We also test SpanGCN using contextualized word embeddings. 
We use ELMo \cite{DBLP:conf/naacl/PetersNIGCLZ18} to train the syntactic parser of \citet{DBLP:conf/acl/KleinK18}, and provide ELMo and RoBERTa \cite{DBLP:journals/corr/abs-1907-11692} embeddings as input to our model.

In Table \ref{tab:conll05-results} (Single / Context. Emb.) we show results of models that employ contextualized embeddings on the CoNLL-2005 test set.
Both SpanGCN models with contextualized embeddings perform better than the models with GloVe embeddings in both test sets.
SpanGCN{\small (RoBERTa)} is outperformed by the syntax-aware model of \citet{DBLP:conf/acl/WangJWSW19} on the WSJ test but obtains results on par with the state of the art \cite{DBLP:journals/corr/abs-1901-05280} on the Brown test set.

When we train the syntax-agnostic baseline of Section \ref{sec:model} with RoBERTa embeddings, we obtain 87.0 F1 on the WSJ test set and 79.7 on the Brown test set, 0.9\% F1 worse than SpanGCN on both test sets. 
This suggests that although contextualized word embeddings contain information about syntax \cite{tenney2018you,DBLP:conf/emnlp/PetersNZY18,hewitt2019structural}, explicitly encoding high-quality syntax is still useful.
SpanGCN{\small (ELMo)} has comparable results to SpanGCN{\small (RoBERTa)} when tested on the WSJ test set, but has a 1.1\% difference when tested on the Brown test set.
This difference is not surprising; BERT-like embeddings have been shown to perform better than ELMo embeddings in various probing tasks \cite{DBLP:conf/naacl/Liu0BPS19}. 
We believe that on top of this, the sheer volume of data used to train RoBERTa (160GB of text) is beneficial in the out-of-domain setting.

We report results with contextualized embeddings on CoNLL-2012 in Table \ref{tab:conll12-results} (Single / Context. Emb.).
SpanGCN{\small (RoBERTa)} obtains the best results.
It is interesting to notice, though, that results of the syntax-aware model of \citet{DBLP:conf/acl/WangJWSW19} are overall (on both CoNLL 2005 - 2012) similar to SpanGCN{\small (RoBERTa)}. 
Also in this setting, SpanGCN{\small (ELMo)} obtains similar (although inferior) results to SpanGCN{\small (RoBERTa)} 86.5 vs. 86.8 F1.
Compared with the best ELMo-based model \cite{DBLP:conf/acl/WangJWSW19}, SpanGCN{\small (ELMo)} obtains similar (0.1\% lower) results.

\begin{table}[t!]
	\center
\scalebox{1}{
	\begin{tabulary}{\columnwidth}{@{}L RR R@{}}
		\toprule
		
		\textbf{Model}
		& $\bm{P}$ & $\bm{R}$ & $\bm{F_1}$\\
		
		\midrule
		
		\small{\citet{YangM17} (\textsc{Seq})}
		& 63.4 & 66.4 & 64.9 \\
		
		\small{\citet{YangM17} (\textsc{All})}
		& 70.2 & 60.2 & 65.5 \\

		\small{\citet{DBLP:conf/emnlp/SwayamdiptaTLZD18}$\dagger\ddagger$}
		& 69.2 & 69.0 & 69.1 \\ 
		\midrule[.03em]	
		
 \small{SpanGCN$\dagger$}
		& 69.8 & 68.8 & 69.3 \\ 
		\bottomrule
	\end{tabulary}
	}
	\caption{Results on FrameNet 1.5 test set using gold frames. $\dagger$ indicates syntactic models and $\ddagger$ indicates multi-task learning models.}
	\label{tab:fn_results}
\end{table}

\paragraph{FrameNet}

On FrameNet data, we compare SpanGCN with the sequential and sequential-span ensemble models of \citet{YangM17}, and with the multi-task learning model of \citet{DBLP:conf/emnlp/SwayamdiptaTLZD18}.
\citet{DBLP:conf/emnlp/SwayamdiptaTLZD18} use a multi-task learning objective where the syntactic scaffolding model and the SRL model share the same sentence encoder and are trained together on disjoint data. 
Like our method, this approach injects syntactic information (though dependency rather than constituent syntax) into word representations used by the SRL model.
We show results obtained on the FrameNet test set in Table \ref{tab:fn_results}.
SpanGCN obtains 69.3\% F1 score.
It performs better than the syntax-agnostic baseline (2.9\% F1) and better than the syntax-agnostic ensemble model (\textsc{ALL}) of \citet{YangM17} (3.8\% F1).
SpanGCN also slightly outperforms (0.2\% F1) the multi-task model of \citet{DBLP:conf/emnlp/SwayamdiptaTLZD18}.

\section{Related Work}
Among earlier approaches to incorporating syntax into neural networks,
\citet{socher-EtAl:2013:EMNLP,tai-socher-manning:2015:ACL-IJCNLP} proposed recursive neural networks that encode constituency trees by recursively creating representations of constituents.
There are two important differences between these approaches and ours.
First, in our model, the syntactic information in the constituents flows back to word representations.
This may be achieved with their inside-outside versions \cite{DBLP:conf/emnlp/LeZ14,DBLP:journals/tacl/TengZ17}. 
Second, these previous models do a global pass over the tree, whereas GCNs consider only small fragments of the graph. This may make GCNs more robust when using noisy, predicted syntactic structures.

In SRL, dependency syntax has gained a lot of attention.
Similarly to this work, \citet{DBLP:conf/emnlp/MarcheggianiT17} encoded dependency structures using GCNs.
\citet{StrubellVAWM18} used a multi-task objective to force the self-attention model to predict syntactic edges. 
\citet{roth-lapata:2016:P16-1} encoded dependency paths between predicates and arguments using an LSTM.
\citet{DBLP:conf/emnlp/LiHCZZLLS18} analysed different ways of encoding syntactic dependencies for dependency-based SRL, while \citet{DBLP:conf/acl/ZhaoHLB18} and \citet{DBLP:conf/emnlp/HeLZ19} proposed an argument pruning technique which calculates promising candidate arguments.
Recently, \citet{DBLP:conf/acl/WangJWSW19} used syntax linearizaton approaches of \citet{gomez-rodriguez-vilares-2018-constituent} and employed this information as a word-level feature in a SRL model.
\citet{DBLP:conf/emnlp/SwayamdiptaTLZD18,DBLP:journals/tacl/CaiL19} used multi-task learning to produce syntactically-informed word representation, with a sentence encoder shared between SRL and an auxiliary syntax-related task.

In earlier work, 
\citet{naradowsky2012} used graphical models to encode syntactic structures while \citet{DBLP:journals/coling/MoschittiPB08} applied tree kernels for encoding constituency trees.
Many methods cast the problem of SRL as a span classification problem.
\citet{fitzgerald-EtAl:2015:EMNLP} used hand-crafted features to represent spans, while \citet{HeLLZ18} and \citet{DBLP:conf/emnlp/OuchiS018} adopted a BiLSTM feature extractor.
In principle, SpanGCN can be used as a syntactic feature extractor within this class of models. 

\section{Conclusions}

In this paper, we introduced SpanGCN, a novel neural architecture for encoding constituency syntax at the word level.
We applied SpanGCN to SRL, on PropBank and FrameNet.
We observed substantial improvements from using constituent syntax on both datasets, and also in the realistic out-of-domain setting. 
By comparing to dependency GCN, we observed
that for SRL constituent structures yield more informative features that the dependency ones.
Given that GCNs over dependency and constituency structure have access to very different information, it would be interesting to see in future work if combining two types of representations can lead to further improvements. 
While we experimented only with constituency syntax, SpanGCN may be able to encode any kind of span structure, for example, co-reference graphs, and can be used to produce linguistically-informed encoders for other NLP tasks rather than only SRL.

\section*{Acknowledgments}
We thank Luheng He for her helpful suggestions. 
The project was supported by the European Research Council (ERC StG BroadSem 678254), and the Dutch National Science Foundation (NWO VIDI 639.022.518).
We thank NVIDIA for donating the GPUs used for this research.

\bibliographystyle{acl_natbib}
\bibliography{emnlp2020}
\clearpage
\newpage

\appendix
\begin{figure}[t]
\begin{center}
\includegraphics[width=1.00\columnwidth]{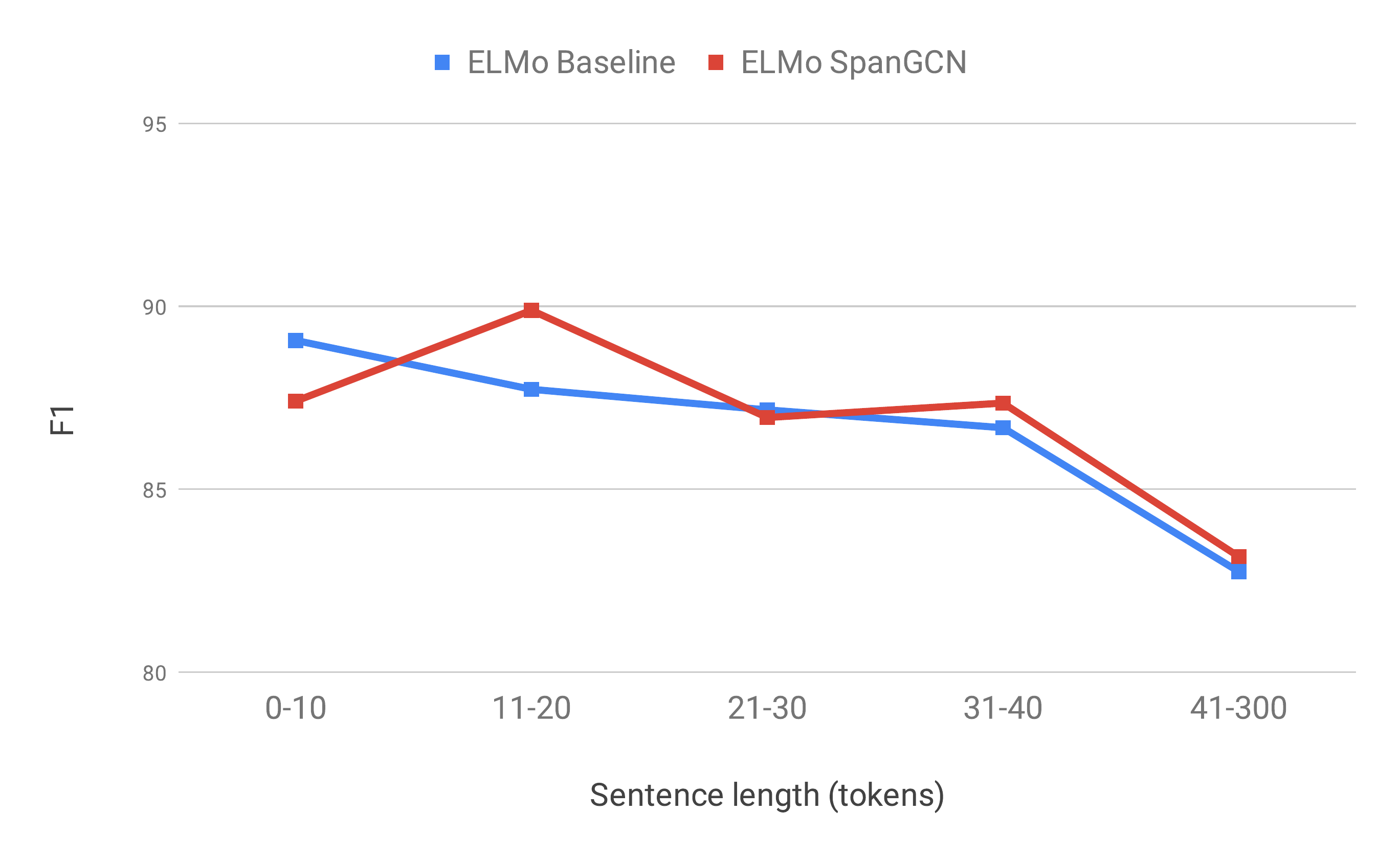}
\includegraphics[width=1.00\columnwidth]{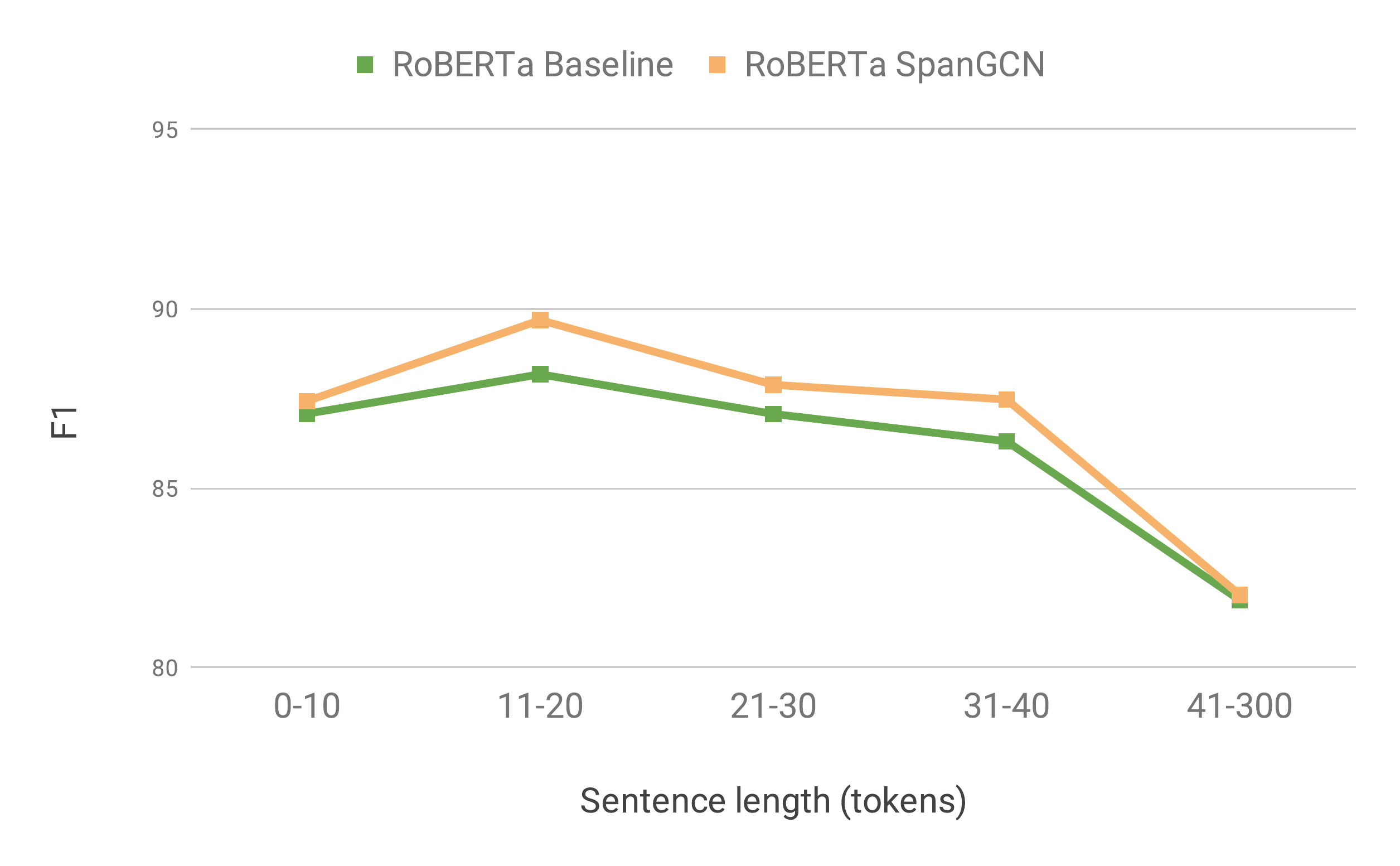}
\caption{CoNLL-2005 F1 score as a function of sentence length.
\label{fig:sent_len_context_embeddings} 
} 
\end{center}
\end{figure}

\begin{figure}[t]
\begin{center}
\includegraphics[width=1.00\columnwidth]{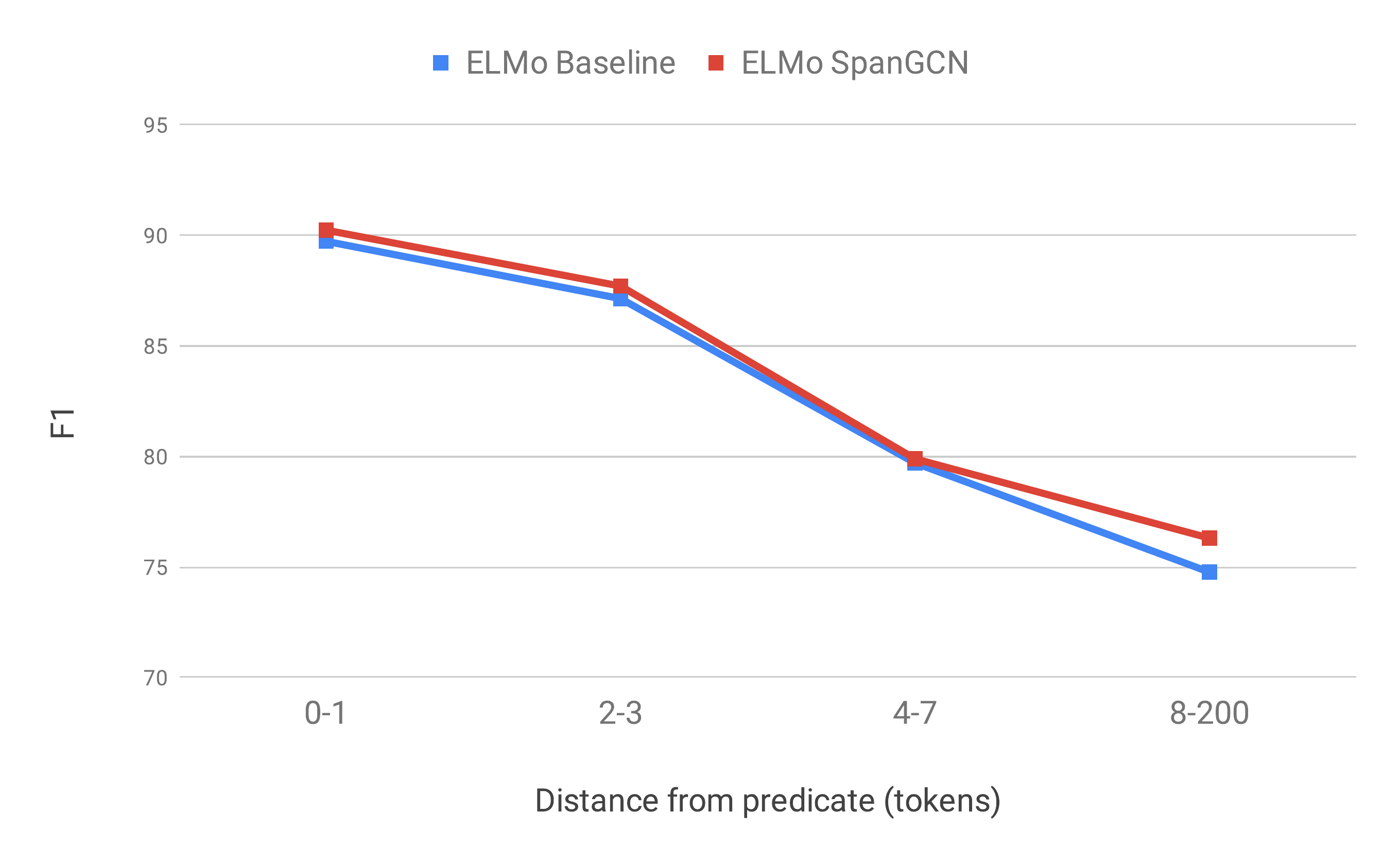}
\includegraphics[width=1.00\columnwidth]{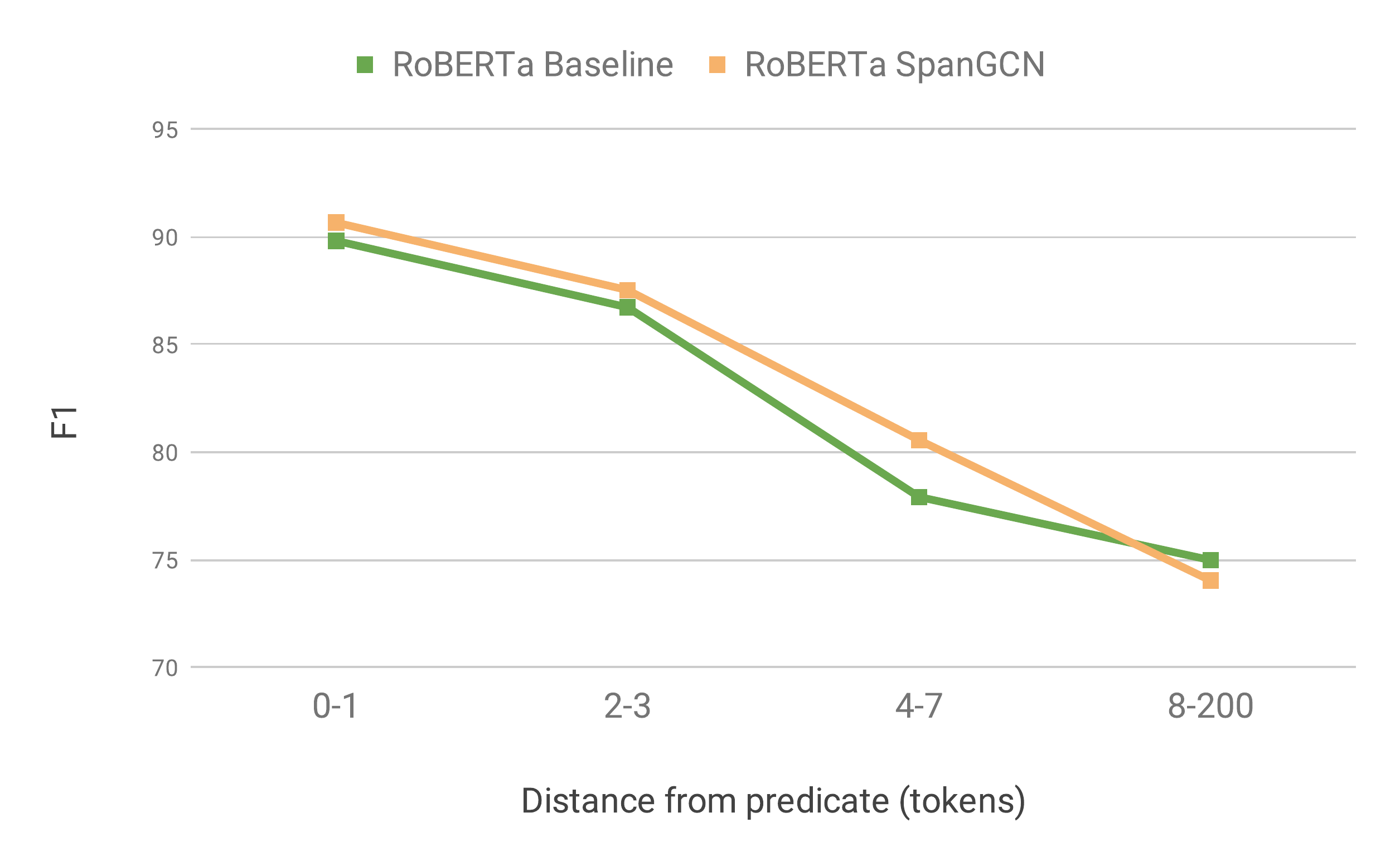}
\caption{CoNLL-2005 F1 score as a function of the distance of a predicate from its arguments.
\label{fig:pred_distance_context_embeddings} 
} 
\end{center}
\end{figure}

\begin{figure}[t]
\begin{center}
\includegraphics[width=1.00\columnwidth]{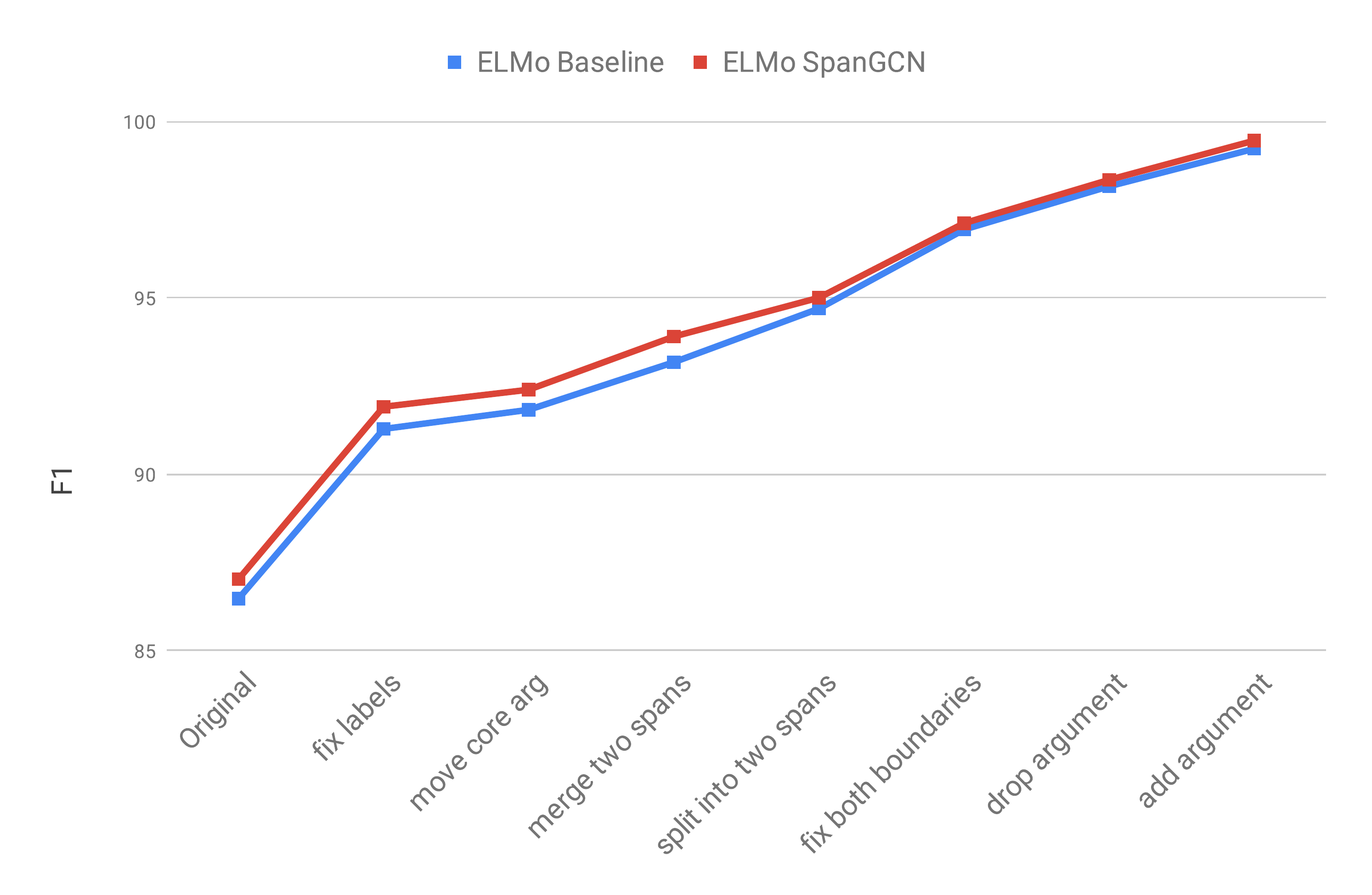}
\includegraphics[width=1.00\columnwidth]{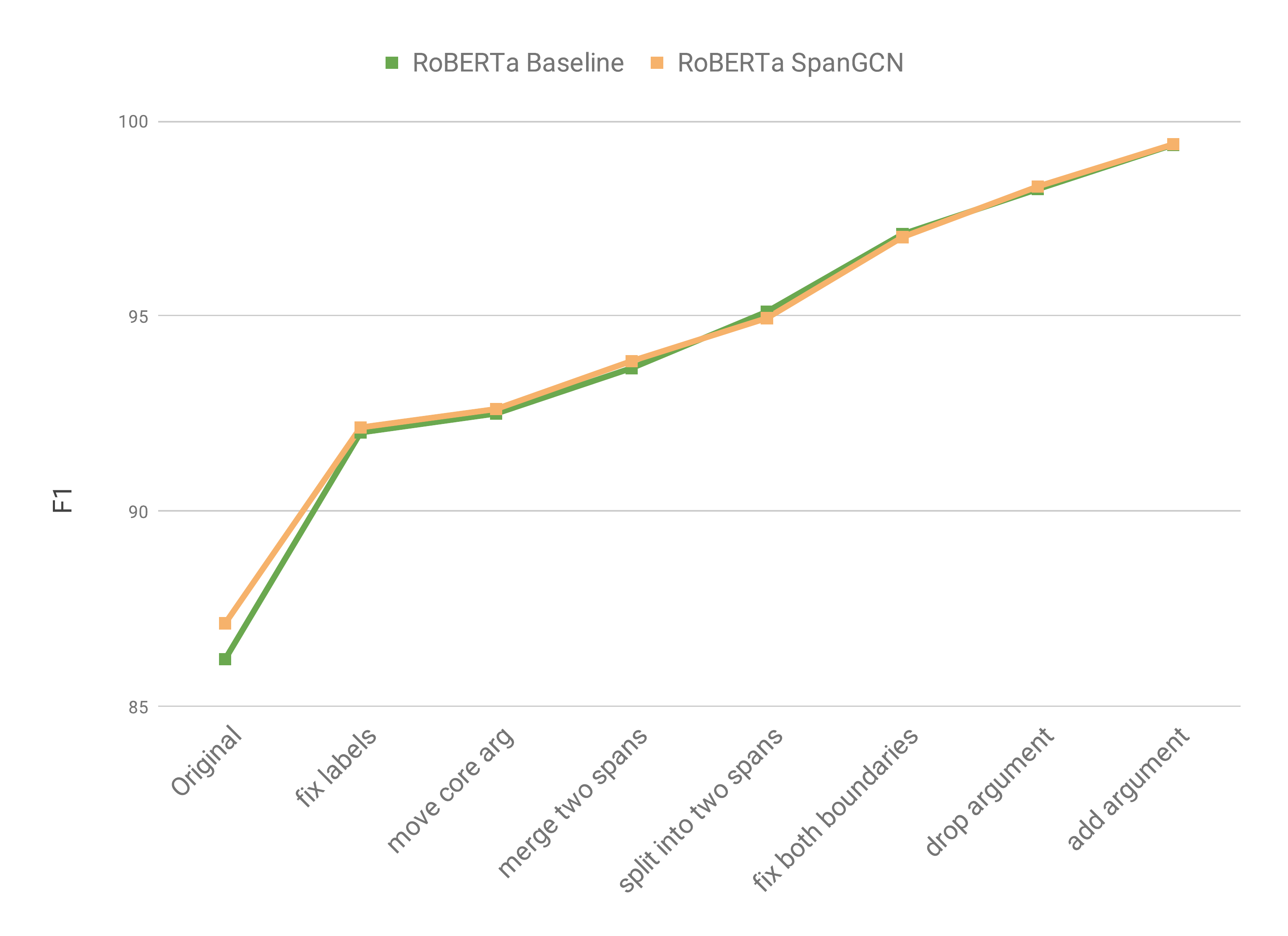}
\caption{Performance of CoNLL-2005 models after performing corrections from \citet{HeLLZ17a}.\label{fig:corrections_context_embeddings} 
} 
\end{center}
\end{figure}

\section{Implementation details}
\label{sec:implementation_details}
We used 100-dimensional GloVe embeddings \cite{pennington-socher-manning:2014:EMNLP2014} for all our experiments unless otherwise specified. 
We tuned the hyperparameters on the CoNLL-2005 development set.
The LSTMs hidden state dimensions were set to 300 for CoNLL experiments and to 200 for FrameNet ones.
In our model, we used a four-layer BiLSTM below GCN layers and a two-layer BiLSTM on top.
We used an eight-layer BiLSTM in our syntax-agnostic baseline; the number of layers was independently tuned on the CoNLL-2005 development set.
For RoBERTa \cite{DBLP:journals/corr/abs-1907-11692} experiments, we used the last layer of the 12-layers (roberta-base) pretrained transformer \cite{DBLP:conf/nips/VaswaniSPUJGKP17} without fine tuning it.
In the case words got split into multiple subwords by the RoBERTa tokenizer, we took the vector of the first subword unit as the representation of the word as in \citet{DBLP:conf/naacl/DevlinCLT19}.

For ELMo experiments, we learned the mixing coefficients of ELMo, and we concatenated the weighted sum of the ELMo layers with a GloVe 100-dimensional vector.
We used the original 5.5B ELMo model \footnote{\url{https://allennlp.org/elmo}}.

For FrameNet experiments, we constrained the CRF layer to accept only BIO tags compatible with the selected frame.

We used Adam \cite{kingma2014adam} as an optimizer with an initial learning rate of 0.001; we halved the learning rate if we did not see an improvement on the development set for two epochs.
We trained the model for a maximum of 100 epochs. 
We clipped the norm of the gradient to 1.

All models were implemented with PyTorch.\footnote{\url{https://pytorch.org}} 
We used some modules from AllenNLP\footnote{\url{https://github.com/allenai/allennlp}} and the reimplementation of the FrameNet evaluation scripts by \citet{DBLP:conf/emnlp/SwayamdiptaTLZD18}.\footnote{\url{https://github.com/swabhs/scaffolding}}

\section{Analysis on Syntax Plus Contextualized Embeddings}

We perform an analysis on the use of syntax on top of contextualized representations ELMo and RoBERTa.
We perform this analysis on the CoNLL-2005 development set and we measure the impact of contextualized syntax-agnostic vs contextualized syntactic model in function of: sentence length (Figure \ref{fig:sent_len_context_embeddings}), of the distance of arguments from the predicate (Figure \ref{fig:pred_distance_context_embeddings}), and in function of the type of mistakes they make (Figure \ref{fig:corrections_context_embeddings}).
In Figures \ref{fig:sent_len_context_embeddings}, \ref{fig:pred_distance_context_embeddings}, and \ref{fig:corrections_context_embeddings}, baselines consist of a syntax agnostic 8 layer BiLSTMs on top of the frozen contextualized representation.

Figure \ref{fig:sent_len_context_embeddings} shows that for both ELMo and RoBERTa, SpanGCN is beneficial. For ELMo though SpanGCN is not helpful for short sentences (up to length 10), while for RoBERTa, the syntax is beneficial across all sentence lengths.

Figure \ref{fig:pred_distance_context_embeddings} shows that syntax is beneficial for both contextualized representations.
An interesting difference is that for ELMo, syntax is more helpful for arguments very far from the predicate. 
In contrast, for RoBERTa, syntax is helpful on arguments 4-7 tokens away from the predicate, but hurts performance on arguments farther away from the predicate.

Finally, in Figure \ref{fig:corrections_context_embeddings}, we see rather different behaviour between the two representations.
For ELMo, the errors that the syntax agnostic model makes are the ones related to span boundaries.
For RoBERTa, the syntax-agnostic model makes errors regarding labels, but it is as good as the syntactic model at predicting span boundaries.

\section{Additional Results}
\label{sec:additional_results}
Additional development results for CoNLL-2005 (Table \ref{tab:conll05-results-app}) and CoNLL-2012 (Table \ref{tab:conll12-results-app}) datasets.
\begin{table*}
\center
\scalebox{0.80}{
\begin{tabular}{llllllllllll}
& \multicolumn{3}{c}{Dev} && \multicolumn{3}{c}{WSJ Test} && \multicolumn{3}{c}{Brown Test} \\ \cline{2-4} \cline{6-8} \cline{10-12}
 Single & $\bm{P}$ & $\bm{R}$ & $\bm{F_1}$ && $\bm{P}$ & $\bm{R}$ & $\bm{F_1}$ && $\bm{P}$ & $\bm{R}$ & $\bm{F_1}$\\ \hline \hline
\citet{HeLLZ17a} & 81.6 & 81.6 & 81.6 & & 83.1 & 83.0 & 83.1 && 72.9 & 71.4 & 72.1\\ 
\citet{HeLLZ18} & - & - & - & & 84.2 & 83.7 & 83.9 && 74.2 & 73.1 & 73.7\\ 
\citet{TanWXCS18} & 82.6 & 83.6 & 83.1 & & 84.5 & 85.2 & 84.8 && 73.5 & 74.6 & 74.1\\
\citet{DBLP:conf/emnlp/OuchiS018} & 83.6 & 81.4 & 82.5 && 84.7 & 82.3 & 83.5 && 76.0& 70.4 & 73.1\\
\citet{StrubellVAWM18}$\dagger\ddagger$ & 83.6 & 83.74 & 83.67 & & 84.72 & 84.57 & 84.64 && 74.77 &74.32 &74.55\\ \hline 
DepGCN$\dagger$ & 83.4 & 83.73 & 83.56 & &85.07& 84.7 & 84.88 && 75.5 & 74.46 & 74.98 \\
SpanGCN$\dagger$ & 84.48 & 84.26 & 84.37 & & 85.8 & 85.05 & {85.43} && 76.17 & 74.74 & 75.45 \\ 
 & & & && & & && & & \\
Single / Contextualized Embeddings & & & && & & && & & \\ \hline \hline
\citet{HeLLZ18}{\small (ELMo)} & - & - & 83.9 & & - & - & 87.4 && - & - & 80.4\\ 
\citet{DBLP:journals/corr/abs-1901-05280}{\small (ELMo)} & - & - & - & & 87.9 & 87.5 & 87.7 && 80.6 & 80.4 & 80.5 \\ 
\citet{DBLP:conf/emnlp/OuchiS018}{\small (ELMo)} & 87.4& 86.3& 86.9 & & 88.2 & 87.0 & 87.6 && 79.9 & 77.5 & 78.7\\
\citet{DBLP:conf/acl/WangJWSW19}{\small (ELMo)}$\dagger$ & -& -& - & & - & - & 88.2 && - & - & 79.3\\
\hline 
Baseline{\small (ELMo)}$\dagger$ & 86.07 & 86.84 & 86.46 & &86.81& 87.13 & 86.97 && 78.43 & 77.81 & 78.12  \\
Baseline{\small (RoBERTa)}$\dagger$ & 85.95 & 86.3 & 86.13  & & 86.85 & 87.19 & 87.02 && 79.99 & 79.33 & 79.66 \\
SpanGCN{\small (ELMo)}$\dagger$ & 86.46 & 87.38 & 86.92 & &87.47& 87.85 & 87.66 && 79.38 & 79.56 & 79.47 \\
SpanGCN{\small (RoBERTa)}$\dagger$ & 86.77 & 87.56 & 87.17 & & 87.72 & 88.05 & 87.89 && 80.45 & 80.71 & 80.58 \\

\hline 
\end{tabular}
}
\caption{Precision, recall and $\bm{F_1}$ on the CoNLL-2005 development and test sets. $\dagger$ indicates syntactic models and $\ddagger$ indicates multi-task learning models. \label{tab:conll05-results-app}}
\end{table*}

\begin{table*}
\center
\scalebox{0.80}{
\begin{tabular}{lllllllll}
& \multicolumn{3}{c}{Dev} && \multicolumn{3}{c}{Test} \\ 
\cline{2-4} \cline{6-8} 
Single & $\bm{P}$ & $\bm{R}$ & $\bm{F_1}$ && $\bm{P}$ & $\bm{R}$ & $\bm{F_1}$ \\ \hline \hline
\citet{HeLLZ17a} & 81.8 & 81.4 & 81.5 & & 81.7 & 81.6 & 81.7 \\ 
\citet{TanWXCS18} & 82.2 & 83.6 & 82.9 & & 81.9 & 83.6 & 82.7 \\
\citet{DBLP:conf/emnlp/OuchiS018} & 84.3 & 81.5 & 82.9 && 84.4 & 81.7 & 83.0 \\
\citet{DBLP:conf/emnlp/SwayamdiptaTLZD18}$\dagger\ddagger$ & - & - & - & & 85.1 & 81.2 & 83.8 \\

\hline 
SpanGCN$\dagger$ & 84.45 & 84.16 & 84.31 & & 84.47 & 84.26 & 84.37 \\ 
 & & & && & & \\

Single / Contextualized Embeddings & & & && & & \\ \hline \hline
\citet{DBLP:conf/naacl/PetersNIGCLZ18}{\small (ELMo)} & - & - & - & & - & - & 84.6 \\
\citet{DBLP:journals/corr/abs-1901-05280}{\small (ELMo)} & - & - & & & 85.7 & 86.3 & 86.0 \\ 
\citet{DBLP:conf/emnlp/OuchiS018}{\small (ELMo)} & 87.2 & 85.5 & 86.3 && 87.1 & 85.3 & 86.2 \\
\citet{DBLP:conf/acl/WangJWSW19}{\small (ELMo)}$\dagger$  & - & - & - && - & - & 86.4\\
\hline 
Baseline{\small (ELMo)}$\dagger$ & 84.55 & 83.7 & 84.13 & & 84.55  & 83.56 & 84.06 \\
Baseline{\small (RoBERTa)}$\dagger$ & 84.6 & 84.69 & 84.64 & & 84.71 & 84.85 & 84.78 \\
SpanGCN{\small (ELMo)}$\dagger$ & 86.26 & 86.74 & 86.5 & & 86.25 & 86.83 & 86.54 \\
SpanGCN{\small (RoBERTa)}$\dagger$ & 86.69 & 87.22 & 86.95 & & 86.48 & 87.09 & 86.78 \\

\hline 
\end{tabular}
}
\caption{Precision, recall and $\bm{F_1}$ on the CoNLL-2012 development and test sets. $\dagger$ indicates syntactic models and $\ddagger$ indicates multi-task learning models. \label{tab:conll12-results-app}}
\end{table*}

\end{document}